\documentclass[submission,copyright,creativecommons]{eptcs}
\providecommand{\event}{AREA 2023} 

\usepackage[ruled,linesnumbered,noend]{algorithm2e}
\usepackage{amsfonts}

\usepackage{iftex}

\ifpdf
  \usepackage{underscore}         
  \usepackage[T1]{fontenc}        
\else
  \usepackage{breakurl}           
\fi

\newcommand{\commentout}[1]{}

\title{Rollout Heuristics for Online Stochastic Contingent Planning}

\author{Oded Blumenthal
\institute{Software and Information Systems Engineering, Ben Gurion University, Israel}
\email{odedblu@post.bgu.ac.il}
\and
Guy Shani
\institute{Software and Information Systems Engineering, Ben Gurion University, Israel}
\email{shanigu@bgu.ac.il}
}
\def\titlerunning{Rollout Heuristics for Online Stochastic Contingent Planning}
\def\authorrunning{O. Blumenthal, G. Shani}

\begin{document}
\maketitle

\begin{abstract}
Partially observable Markov decision processes (POMDP) are a useful model for decision-making under partial observability and stochastic actions. Partially Observable Monte-Carlo Planning is an online algorithm for deciding on the next action to perform, using a Monte-Carlo tree search approach, based on the UCT (UCB applied to trees) algorithm for fully observable Markov-decision processes. POMCP develops an action-observation tree, and at the leaves, uses a rollout policy to provide a value estimate for the leaf. As such, POMCP is highly dependent on the rollout policy to compute good estimates, and hence identify good actions. Thus, many practitioners who use POMCP are required to create strong, domain-specific heuristics.

In this paper, we model POMDPs as stochastic contingent planning problems. This allows us to leverage domain-independent heuristics that were developed in the planning community. We suggest two heuristics, the first is based on the well-known $h_{add}$ heuristic from classical planning, and the second is computed in belief space, taking the value of information into account.

\end{abstract}

\section{Introduction}

Many autonomous agents operate in environments where actions have stochastic effects, and important information that is required for obtaining the goal is hidden from the agent. Agents in such environments typically execute actions and sense some observations that result from these actions. Based on the accumulated observations the agents can better estimate their current state and decide on the next action to execute.
Such environments are often modeled as partially observable Markov decision processes (POMDPs) \cite{smallwood1973optimal}. 

POMPD models allow us to reason about the hidden state of the system, typically using a belief state -- a distribution over the possible environment states. The belief state can be updated given the executed action and the received observation. One can compute a policy, a mapping from beliefs to actions, that dictates which action should be executed given the current belief. Many algorithms were suggested for computing such policies \cite{shani2013survey}.

However, in larger environments, it often becomes difficult to maintain a belief state, let alone compute a policy for all possible belief states. In such cases, one can use an online re-planning approach, where after every action is executed, the agent computes which action to execute next. Such online approaches replace the lengthy single policy computation which is done offline, before the agent begins to act, with a sequence of shorter computations, which are executed online, during execution, after each action \cite{ross2008online}.

POMCP \cite{silver2010monte} is such an online replanning approach, extending the UCT algorithm for fully observable Markov decision processes (MDPs) to POMDPs. POMCP operates by constructing online a search tree, interleaving decision and observation nodes. The root of the tree is a decision node. Each decision node has an outgoing edge for every possible action, ending at an observation node. Then, the outgoing edges from an observation node denote the possible observations that result from the incoming action. The agent computes a value for each node in the tree, and then, the agent can choose, from the root node, the action associated with the edge leading to the highest value child node.

To evaluate the value of leaf nodes, POMCP executes a random walk in belief space, known as a rollout, where the agent selects actions from some rollout policy to construct a trajectory in belief space and obtain an estimation of the quality of the leaf node. Clearly, this evaluation is highly dependant on the ability of the rollout policy to reach the agent goals. In complex problems, obtaining the goal may require a lengthy sequence of actions \cite{kurniawati2022partially}, and until the goal is reached, no meaningful rewards are  obtained. Indeed, practitioners that use POMCP often implement complex domain-specific heuristics for the rollout policy.

In this paper we focus on suggesting domain-independent heuristics for rollout policies. We leverage work in automated planning, using heuristics defined for classical and contingent planning problems \cite{helmert2009landmarks,helmert2014merge,geffner2022concise}. We thus represent POMDP problems in a structured manner, using boolean facts to capture the state of the environments. This allows both for a compact representation of large problems, compared with standard flat representations that do not scale, as well as the ability to use classical planning heuristics.

We begin by suggesting using the well-known $h_{add}$ heuristic for choosing rollout actions \cite{bonet2001planning}. This heuristic searches forward in a delete relaxation setting, until the goal has been reached. Then, the value of an action is determined by the number of steps in the delete relaxation space following the action, required for obtaining the goal.

Next, we observe that any state-based rollout policy is inherently limited in its ability to evaluate the missing information required for reaching the goal, and hence, provide some estimate as to the value of information \cite{howard1966information} of an action. We hence suggest a multi-state rollout policy, where actions are executed on a set of states jointly, and observations are used to eliminate states that are incompatible with the observed value. We show that this heuristic is much more informed in domains that require complex information-gathering strategies.

For an empirical evaluation, we extend domains from the contingent planning community with stochastic effects. We evaluate our heuristics, comparing them to random rollouts, showing that they allow us to provide significantly better behavior.

\section{Background}

We now provide the required background on POMDPs, contingent planning, domain-independent heuristics, and the POMCP algorithm.

\subsection{POMDPs}

A goal-oriented partially observable Markov decision process (POMDP) 
is a tuple $\langle S,A,tr,\Omega,O,G \rangle$ \cite{bonet2009solving}. $S$ is a set of states; $A$ is a set of actions. $tr:S \times A \times S \rightarrow [0,1]$ is the transition function, i.e., $tr(s,a,s')$ is the probability that when executing action $a$ at state $s$ we would reach state $s'$.  $\Omega$ is the set of observations the agent can obtain. $O:S \times A \times \Omega \rightarrow [0,1]$ is the observation function, such that $O(s,a,o)$ is the probability of observing $o$ when $a$ was performed and led to state $s$. $G$ is a set of goal states.

Because the state of a POMDP is partially observable, the agent typically does not know what the true underlying state of the world is.
Hence, it can maintain a {\em belief state\/} $b$, which is a distribution over $S$, i.e., $b(s)$ is the likelihood that $s$ is the current state. When a goal belief is reached, that is $sum_{s\in G}b(s)=1$, then the agent is sure that it is at a goal state, and the execution terminates.

A solution to a POMDP, called a \emph{policy}, is a function that assigns an action to every belief state. The optimal policy minimizes the expected cost, i.e., the expected number of steps before a goal belief has been reached.

\subsection{POMCP}

The Partially Observable Monte-Carlo Planning (POMCP) online re-planning approach uses a Monte Carlo tree search (MCTS) approach to select the next action to execute \cite{silver2010monte}. At each re-planning episode POMCP constructs a tree, where the root node is the current belief state. Then, POMCP runs forward simulations, where a state is sampled from the current belief, and actions are chosen using an exploration strategy that initially selects actions that were not executed a sufficient amount of times, but gradually moves to select the seemingly best action. Observations in the simulations are selected based on the current state-action observation distribution.

When reaching a leaf node, POMCP begins a so-called rollout. This rollout is designed to provide a value estimate for the leaf, based on some predefined rollout policy. The value of the leaf is updated using the outcome of the rollout, and then the values of the nodes along the branch that were visited during the simulation are updated given their descendants. Obviously, the values of all nodes in the tree are hence highly dependent on the values obtained by the rollout policy.

POMCP is an anytime algorithm, that is, it continues to run simulations until a timeout, and then returns the action that seems best at the root of the search tree.

POMCP was designed for large problems. Hence, POMCP does not maintain and update a belief state explicitly. Instead, POMCP uses a particle filter approach, where a set of states is sampled at the initial belief, and this set is progressed through the tree.

\subsection{Contingent Planning under Partial Observability}

A partially observable contingent planning problem is a tuple: $\pi=\langle P,A_{act},A_{sense},\varphi_I,G
\rangle$ \cite{hoffmann2005contingent,albore2009translation,bonet2011planning,shani2011replanning}. $P$ is a set of facts, $A_{act}$ is a set of actuation actions, and $A_{sense}$ is a set of sensing actions. $\varphi_I$ is a formula describing the set of initially possible states. For ease of exposition, we will assume that $\varphi_I$ is a conjunction of facts, disjunctions of facts, or {\em oneof} clauses over facts, specifying that exactly one fact in the clause holds. 
A state $s$ assigns truth values to all $p \in P$.
$G$ is a formula over $P$ defining goal conditions.

A \emph{belief-state} is a set of possible states. The initial belief
state, $b_I = \{s : s\models\varphi_I\}$ is the set of initially possible states. 

An actuation action $a\in A_{act}$ is a pair,
\{\emph{pre}($a$), \emph{eff}($a$)\}, where 
\emph{pre}($a$) is a set of fact preconditions, and
\emph{eff}($a$) is a set of pairs $(c,e)$ denoting conditional effects. 
We use $a(s)$ to denote the state that is obtained when $a$ is executed in state $s$. 
A sensing action $a\in A_{sense}$ is a pair,
\{\emph{pre}($a$), \emph{obs}($a$)\}, where 
\emph{pre}($a$) is as above, and \emph{obs}($a$) is a set of facts in $P$ whose value is observed when $a$ is executed. We denote by $obs(a,s)$ the values of the observed facts when $a$ is executed at state $s$.
This separation to actuation and sensing actions is only for ease of exposition, and our methods apply also to actions that both modify the state of the world and provide an observation.

Preconditions allow us to restrict our attention only to applicable actions. An action is applicable in a belief $b$ is all possible states in $b$ satisfy the preconditions of the action. Obviously, one can avoid specifying preconditions for actions, allowing for actions that can always be executed, as is typically the case in flat POMDP representations. However, in domains with many actions, preconditions are a useful tool for drastically limiting the amount of actions that should be considered.

\subsection{Regression-based Belief Maintenance}

Updating a belief can be costly. Alternatively, one can avoid the computation of new formulas representing the updated belief, by maintaining only the initial belief formula, and the history --- the sequence of executed actions and sensed observations \cite{brafman2016online}. When the agent needs to query whether the preconditions of an action or the goal hold at the current node, the formula is regressed \cite{rintanen2008regression} through the action-observation sequence back towards the initial belief. Then, one can apply SAT queries to check whether the query formula holds. We now briefly review the regression process for deterministic actions. This approach can be highly useful for larger POMDPs, complementing the particle filter approach used in POMCP.

First, let us consider the regression of an actuation action $a$ that does not provide an observation.
Let $\phi$ be a propositional formula and $a$ a \emph{deterministic} actuation action. 
Let $c_{a,l}$ denote the condition under which $l$ is an effect of $a$, and that $a(s)$ satisfies $l$ iff either $s\models c_{a,l}$ or $s\models l\wedge\neg c_{a,\neg l}$. Hence,
we define the \emph{regression} of $\phi$ with respect to $a$ as:
\begin{eqnarray}
&rg_a(\phi) = pre(a) \wedge \phi_{r(a)}& \label{eqn:Regression1}\\
&\phi_{r(a)} = \mbox{replace each literal $l$ in $\phi$ by } c_{a,l} \vee (l \wedge \neg c_{a,\neg l})&\label{eqn:Regression2}
\end{eqnarray}

Now, consider a sensing action and an ensuing observation. 
Suppose we want to validate that $\phi$ holds following the execution of $a \in A_{sense}$ in some state $s$  given that we observed $obs(a)=o$. Thus, we need to ensure that following $a$, if $l$ holds then $\phi$ holds. That is:
\begin{equation}
rg_{a,o}(\phi) = rg_a(obs(a)=o\rightarrow\phi)\label{eqn:Regression3}
\end{equation}

Regression maintains the equivalence of the formula \cite{rintanen2008regression,brafman2016online}.
For any two formulas $\phi_1$ and $\phi_2$ we have:
\begin{enumerate}
\item $\phi_1\equiv\phi_2 \Rightarrow rg_{a,o}(\phi_1) \equiv rg_{a,o}(\phi_2)$
\item $\phi_1\equiv\phi_2 \Rightarrow rg_{a}(\phi_1) \equiv rg_{a}(\phi_2)$
\end{enumerate}
Hence, we can produce a regression over formulas, and compare the regressed formulas, making conclusions about the original formulas.

The regression can be recursively applied to a sequence of actions and observations (history) $h$ as follows:
\begin{equation}
rg_{h+(a,o)}(\phi) = rg_{h}(rg_{a,o}(\phi));\ \ \ rg_{\epsilon,\epsilon}(\phi) = \phi \label{eqn:Regression4}
\end{equation}
where $\epsilon$ is the empty sequence. This allows us to perform a regression of a formula through an entire plan, and analyze the required conditions for a plan to apply.

In addition, the regression mechanism of \cite{brafman2016online} maintains a cached list of fluents $F(n)$ that are known to hold at node $n$, given the action effects or observations. 
Following an actuation action $a$, $F(a(n))$ contains all fluents in $F(n)$ that were not modified by $a$, as well as effects of $a$ that are not conditioned on hidden fluents. For a sensing action revealing the value $l$, $F(a(n,l)) = F(n) \cup \{l\}$. During future regression queries, when a value at a particular node becomes known, e.g. when regressing a later observation, it is added to $F(n)$.
All fluents $p$ such that $p \notin F(n) \wedge \neg p \notin F(n)$ are said to be hidden at $n$. The cached list is useful for simplifying future regressed formulas.

\section{Related Work}

Augmenting MCTS methods with heuristics in the context of fully observable MDPs was previously suggested. \cite{keller2013trial} describe an MCTS tree based approach that uses planning based heuristics. The PROST planner \cite{keller2012prost} uses a heuristic for estimating the value of states. The DP-UCT approach \cite{shen2019guiding} use a planning heuristic based on deep learning for the rollout phase. We are not aware of previous attempts to adapt these approaches to POMDPs.

There were several extensions suggested for POMCP \cite{kurniawati2022partially}. For example \cite{kurniawati2016online} considers dynamic environments, and \cite{hoerger2019multilevel} consider non-linear dynamics. All these methods rely on rollouts, and can hence leverage the rollout strategies that we suggest here.

POMCPOW \cite{sunberg2018online} extends POMCP to the challenge of solving POMDPs with continuous state, action, and observation spaces. It constructs the search tree incrementally to explore additional regions of the observation and action spaces. It also requires a rollout policy to evaluate the utility of leaf nodes.
Our rollout strategies relay on classical planning approaches which are discrete, and it is hence unlikely that our methods can be extended to continuous domains.

DESPOT \cite{somani2013despot,ye2017despot,luo2019importance} is an online POMDP solver based on tree search, similar to POMCP. DESPOT uses a different strategy than the UCB rule for constructing the tree, designed to avoid the overly greedy nature of POMCP exploration strategy. DESPOT also requires a so-called default policy to evaluate the utility of a leaf in the tree, and the authors stress the importance of a strong default policy to improve the convergence. Thus, our methods can be directly applied to DESPOT as well. We chose here to focus on POMCP rather than DESPOT, because POMCP is a simpler method, which allows us to better focus on the importance of the heuristic function, independent of the effect of the various augmentations that DESPOT adds on top of POMCP.

\commentout{
DESPOT enjoys the same strengths as POMCP—breaking the two curses through sampling—but avoids POMCP’s extremely poor worst-case behavior by evaluating policies on a small number of sampled scenarios \cite{ng2013pegasus}.

In each planning step, the algorithm searches for a good policy derived from a Determined Sparse Partially Observable Tree (DESPOT) for the current belief, and executes the policy for one step. A
DESPOT summarizes the execution of all policies under K-sampled scenarios. It is structurally
similar to a standard belief tree, but contains only belief nodes reachable under the K scenarios. leading to a dramatic improvement in computational efficiency
when K is small \cite{somani2013despot}. In our paper, we also use sub-sampling from the belief state, but in different from DESPOT we are using a larger K value( e.g 500~).

Regularized DESPOT (R-DESPOT), which interprets
this lower bound as a regularized utility function, which it uses to optimally balance the size of a
policy and its estimated performance under the sampled scenarios\cite{somani2013despot}. In contrast to R-DESPOT we are not dynamically change the size of the policy and its set to constant value.

\subsection{Importance sampling for online planning under uncertainty}
Another article that extended DESPOT, created an algorithm called importance sampling DESPOT (IS-DESPOT). In order to improve the performance in cases of uncertainty. By using importance sampling on DESPOT, performance on critical but rare events, which are difficult to sample, increases.\cite{luo2019importance}
}

\cite{saborio2019planning} suggest a method called PSG to evaluate the proximity of states to the goal. They suggest using PSG in several places within POMCP including rollouts. PSG assumes that states are defined in a factored manner using state features, and computes a function from features to the goal using subgoals. In essence, their approach can be considered as a type of heuristic, which is highly related to the concept of landmarks in classical planning \cite{richter2008landmarks}. Representing the POMDP as a stochastic contingent planning problem, as we do, allows us to use any heuristic developed in the planning community, and can hence be considered to be an extension of PSG.

\commentout{
In this paper, the researchers used a method called incremental refinement (IRE) to give each feature-action pair a relevance score. Next, they use these scores to prune branches that are less than a threshold value, by this pruning they decrease the average branching factor for each step \cite{saborio2020efficient}. With that being said our algorithm deals with the branching-factor complexity problem by applying max tree expansion, and heuristic rollouts for choosing actions with more meaningful rewards. 
}

Similarly, \cite{xiao2019online} also find it difficult to provide good rollout strategies to compute the value of POMCP leaves. Focusing on a robotics motion planning domain, they suggest SVE, state value estimator, that attempts to evaluate the utility of a state directly.

\commentout{
\subsection{An online POMDP solver for uncertainty planning in dynamic environment}
In this paper, the writers propose a new online POMDP solver, called
Adaptive Belief Tree (ABT), that can reuse and improve existing solution, and update the solution as needed whenever the POMDP model changes \cite{kurniawati2016online}. This has a lot of similarities for our heuristic-guild POMCP algorithm, as when we take a step, we use our previous rollouts and node values and use them for the next step of the algorithm. In contrast to this article, our algorithm is not focused on the changes between policies, which can be very hard in complex large-sized domains.
}

\cite{kim2019pomhdp} also focus on the need to use heuristics for guiding search in POMDPs. They focus on RTDP-BEL \cite{bonet2011planning}, an algorithm that runs forward trajectories in belief space to produce a policy. They show that using a heuristic can significantly improve RTDP-BEL. They use domain specific heuristics, and as such, our domain independent approach can also be applied to their methods.

\section{POMCP for Stochastic Contingent Planning}

We focus here on goal-oriented POMDP domains specified as stochastic contingent planning problems. We now define this concept formally.

We define a stochastic formula $\psi$ to be a set of options. Each option $o$ is a conjunction of facts, and is associated with a probability $pr(o)\in(0,1)$ such that $\sum_o pr(o) =1$. One can sample a single option from the stochastic formula, given the distribution defined by $pr(o)$. 

A stochastic contingent planning problem is a tuple $\pi=\langle P,A_{act},A_{sense},\varphi_I, pr_I,G
\rangle$, where $P, A_{sense}$, $\varphi_I, G$ are as in a deterministic contingent planning problem.  $pr_I$ is a stochastic formula defining probabilities over the initial values of some unknown facts. For each action $a\in A_{act}$, the formula defining the effects of $a$ may contain stochastic formulas, capturing stochastic effects. We denote by $a(s)$ the distribution over next states given that $a$ was executed at $s$.

This definition does not support noisy observations, however, this is not truly a limitation. One can compile a noisy observation into a deterministic observation over an artificial fact whose value changes stochastically. Consider, for example, a sensor that nosily detects whether there is a wall in front of a robot. Instead of noisily observing whether there is a wall, we can deterministically detect a green light that is lit when the sensor (stochastically) detects a wall. That is, we can observe the green light without noise, but the green light is only noisily correlated with the existence of a wall.

\begin{algorithm}[t!]
\caption{POMCP for Stochastic Contingent Problems}
\label{alg:POMCP}

\SetKwBlock{PS}{Search($h$)}{end}
\SetKwBlock{PSI}{Simulate($s,n,depth$)}{end}
\SetKwBlock{PR}{Rollout($s,B$)}{end}

\PS{
    \While{timeout not reached}{
        $s \sim \varphi_I, pr_I$\\
        $s' \leftarrow $ apply $h$ to $s$\\
        $Simulate(s',root,o)$
     }
     
 }
 
 \PSI{
   
    add $s$ to $n.b$ \\    
    $count(n) \leftarrow count(n)+1$\\
    \If{$G$ is satisfied in $n.history$}{
        $V(n)\leftarrow 0$\\
        \Return
    }
    \If{$depth > MaxTreeDepth $}{
        $V(n) += Rollout(s,n.b)$\\
         \Return
    }
    \If{$n$ is a leaf node}{
        \For{$a \in A_{act} \cup A_{sense}$}
        {
            \If{$pre(a)$ are satisfied at $n.history$}{
                Add child $n.a$ to $n$\\
                \For{$o \in obs(a)$}{
                    Add child $n.a.o$ to $n.a$\\
                }
            }
            
        }
    }
    $a \leftarrow argmin_a Q(n,a) - c \sqrt{\frac{log(count(n))}{count(n.a)}}$\\
    \If{$a \in A_{act}$}{
        $s' \sim a(s)$, $o \leftarrow null$
    }
    \Else{
        $s' \leftarrow s$, $o \leftarrow obs(a,s)$
    }
    $Simulate(s',n.a.o, depth + 1)$\\
    $count(n.a) \leftarrow count(n.a)+1$, $count(n.a.o) \leftarrow count(n.a.o)+1$\\
    $V(n.a) \leftarrow \frac{\sum count(n.a.o) \cdot V(n.a.o)}{count(n.a)}$\\
    $V(n) \leftarrow \min_a V(n.a)$
 }
\PR{
    $depth \leftarrow 0$\\
    \While{$s \notin G \wedge depth < MaxRolloutDepth$}{
        $a \leftarrow \pi_{rollout,B}(s)$\\
        $s \sim a(s)$\\
        \If{$a$ is a sensing action}{
            Remove from $B$ states that do not agree with $s$ on the observation
        }
        $depth \leftarrow depth + 1$
    }
    \Return{$depth$}
}
 
 \end{algorithm}

Algorithm~\ref{alg:POMCP} describes the POMCP implementation for stochastic contingent planning problems. When the agent needs to act, it calls Search. 
Search repeatedly samples a state (line 3-4) and simulates forward execution given this state is the true underlying system state.

We do not maintain or update a belief state. Instead, we use regression over the history of executed actions and sensed observations. The Search procedure hence samples a state $s$ from the initial belief state, given the initial probability distribution $pr_I$ (line 3). Then, the agent advances the sampled state through the history to obtain a current state $s'$ (line 4).

The Simulate procedure is recursive. We first check whether the current tree node is a goal belief. This is done be regressing the negation of the goal formula $\neg G$ through the history of the current node. For goal beliefs, the value is 0, and we can stop.

Our implementation of POMCP also stops deepening the tree after a predefined threshold Max-Tree-Depth. If that threshold is reached (line 12), we run a Rollout to compute an estimation for the cost of reaching the goal from this node (line 13).

In line 15 we check whether this node has already been expanded, and if not, we compute its children. We do so only for applicable actions whose preconditions are satisfied in the current belief (line 17). Again, this is computed using regression over the history. 

We now select an action $a$ using the UCT exploration-exploitation criterion (line 21), and sample a next state and an observation (lines 22-27). We call Simulate recursively in line 28.

Lines 29-32 update the value of the current node. Lines 28,29 update the counters for the executed action and received observation. Line 31 computes the value for the action as a weighted average over all observations. Line 32 computes the value of the node as the minimal cost among all actions. Our value update, which we empirically found to be more useful, is different than the original POMCP, which uses incremental updates, and more similar to the value update in DESPOT \cite{luo2019importance}.

The Rollout procedure receives as input the current simulated state $s$, as well as a set $B$ of states (particles) in the node from which the rollout begins. $B$ is used by some of our rollout heuristics, as we explain below. The rollout executes actions given the heuristic rollout policy until the goal has been reached, or a maximal number of steps has been reached.

\section{Domain Independent Heuristics for POMCP}

We now describe the main contribution of this paper --- two domain independent rollout heuristics that leverage methods developed in the automated planning community, using the structure specified in the stochastic contingent planning problem.

\subsection{Delete Relaxation Heuristics}

Delete relaxation heuristics are built upon the notion that if actions have only positive effects, then the number of actions that can be executed before the state becomes fixed is finite, and in many cases, small. Also, as actions cannot destroy the precondition of other actions, one can execute actions in parallel. Algorithm~\ref{alg:HS} portrays a delete relaxation heuristic.

Delete relaxation heuristics create a layered graph, interleaving action and fact layers. The first layer, which is a fact layer, contains all the facts that hold in the state for which the heuristic is computed (line 2). The second layer, which is an action layer, contains all the actions whose preconditions hold given the facts in the first layer (line 5). The next layer, which is again a fact layer, contains all the positive effects of the actions in the previous layer, as well as all facts from the previous layer (line 6), and so forth. We stop developing the graph once no new facts can be obtained (line 8).

After the graph is created, one can compute a number of heuristic estimates. The $h_{max}$ returns the depth of the first fact layer that satisfies $G$.
The $h_{add}$ heuristic sums the fact depth of all goal facts (line 9) \cite{bonet2001planning}. The $h_{ff}$ heuristic computes a plan in the relaxed space by tracing back actions that achieved the goal predicates \cite{hff}.

In this paper we experimented using the $h_{add}$ heuristic.

\begin{algorithm}[t!]
\footnotesize
\caption{Single State $h_{add}$}
\label{alg:HS}

\SetKwBlock{HADD}{$\pi_{h_{add}}(s)$}{end}

\HADD{
    $fact_0 \leftarrow $ all facts in $s$\\
    $i\leftarrow 1$\\
    \Repeat{$fact_{i-1} = fact_{i-2}$}{
        $action_i \leftarrow \{a \in A_{act} : fact_{i-1} \models pre(a), a \notin \bigcup_{j=1..i-1} action_j$\}\\
        $fact_i \leftarrow fact_{i-1} \cup \{f \in eff(a): a \in action_i\}$\\
        $i \leftarrow i + 1$\\
    }
    \Return{$\sum_{f \in G} i : f \in fact_i, f \notin fact_{i-1}$}
}

 \end{algorithm}

\begin{algorithm}[t!]
\footnotesize
\caption{Belief Space $h_{add}$}
\label{alg:HB}


\SetKwBlock{HSDR}{$\pi_{h_{add}}(s,B)$}{end}

\HSDR{
    $\forall s' \in B, fact^{s'}_0 \leftarrow $ all facts in $s'$\\
    $B_0 \leftarrow B$\\
    $i\leftarrow 1$\\
    \Repeat{$B_{i-1} = B_{i-2} \wedge  \forall s' \in B_{i-1}: fact^{s'}_{i-1} = fact^{s'}_{i-2}$}{
    $B_i \leftarrow B_{i-1}$\\
       $action_i \leftarrow \{a \in A_{act} : \forall s' \in B_i, fact^{s'}_{i-1} \models pre(a), a \notin \bigcup_{j=1..i-1} action_j$\}\\
        \For{$a \in A_{sense}$}{
            \If{$\forall s' \in B_i, fact^{s'}_{i-1} \models pre(a)$}{
                $F \leftarrow $ the values of $obs(a)$ in $fact^s_i$\\
                \For{$s' \in B_i, s' \neq s$}{
                    $F' \leftarrow $ the values of $obs(a)$ in $fact^{s'}_i$\\
                    \If{$F' \neq F$}{
                        $B_i \leftarrow B_i \setminus \{s'\}$\\
                    }
                }
            }
        }
        
        $\forall s' \in B_i, fact^{s'}_i \leftarrow fact^{s'}_{i-1} \cup \{f \in eff(a): a \in action_i\}$\\
        $i \leftarrow i + 1$\\
    }
    \Return{$\sum_{f \in G} i : f \in fact^s_i, f \notin fact^s_{i-1}$}
}

 \end{algorithm}

\subsection{Heuristics in Belief Space}

A major disadvantage of the above heuristics is that they focus on a single state. When the agent is aware of the true state of the system, observations have no value. Hence, the above heuristics, as well as any heuristic that is based on a single state, do not provide an estimate for the value of information, which is a key advantage of POMDPs. We hence suggest now a heuristic that is computed over a set $B$ of possible states (Algorithm~\ref{alg:HB}).

We compute again the delete relaxation graph, with a few modifications. We compute for each state in $B$ a separate fact layer. An action can be applied only if its preconditions are satisfied in the fact layers of all agents (line 7). This is equivalent to the requirement in contingent planning where an action is applicable only if it is applicable in all states in the current belief, where $B$ is served as an approximation of the true belief state.

Second, our method leverages the deterministic observations, that allow us to filter out states that are inconsistent with the received observation (lines 8-14). When a sensing action can be applied, all states that do not agree with the value of $s$ on the observation are discarded from $B$ (lines 10-14). That is, we remove the fact layers corresponding to these states, and no longer consider them when computing which actions can be applied.

We stop when both no states were discarded, and no new facts were obtained (line 17).
This process must take into account sensing actions to remove states that are incompatible with $s$, which would allow, at the next iteration, that action preconditions would be satisfied for less states, and hence additional actions can be executed.

\section{Empirical Evaluation}

We conduct an empirical study to evaluate our methods. Our methods are implemented in C\#. 

\subsection{Benchmark Domains}

We extended the following contingent planning benchmarks to stochastic settings:

\paragraph{Doors:} In the door domain the agent must move in a grid to reach a target position. Odd levels in the grid are all open, while in even levels there are doors, and only one door is open. The agent can sense whether a door is open when it is at adjacent cells. The agent must identify the open doors and get to the target position. In the stochastic version the agent can open a closed door with some probability of success. The agent can hence either search for the already open door, or attempt to open a closed door.

\paragraph{Blocks World:} In the contingent blocks world problem, the agent does not know the structure of the initial block configuration, but it can sense whether one block is on top of another one, and whether a block is clear. In the stochastic version moving a block from one block to another has a 0.3 probability of success, while moving blocks to and from the table succeeds deterministicly. Hence, it is often preferable to use the table as an intermediate position.

\paragraph{Unix:} In this domain the agent must search for a file in a file system, and copy it to a destination folder. In the stochastic version there is a non-uniform distribution over the possible locations of the file.

\paragraph{MedPks:} The agent here needs to identify which illness a patient has and treat it. To do so, the agent tests for each illness independently, until the proper illness is found. The stochastic version here has non-uniform distribution over the possible illnesses as well.

\paragraph{Localize:} In this domain the agent must reach a goal position in a grid. The agent does not know where it initially is, and can only sense adjacent walls. In the stochastic version there several places in the grid where the agent may slip and stay in place. This makes the localization in the grid more difficult. 

\paragraph{Wumpus:} In this challenging problem the agent must reach a target position in a grid infested by monsters called Wumpuses. Cells may be unsafe to travel as they may contain either a Wumpus or a pit. Wumpuses emit a stench, and pits emit a breeze, both of which can be sensed in adjacent cells. The agent must sense in multiple positions to identify the safe cells. The stochastic version here also has non-uniform distribution over the safe cells.

\subsection{Results}

For each domain above we run 20 online episodes, and compute the success rate, the average run time for computing the next action, and the average cost to the goal. We did not use a timeout, but runs longer than 100 steps were considered to be stuck in a loop, and terminated.

\begin{table}[t]
\footnotesize
    \begin{tabular}{|l|c|c|c|c|c|c|c|c|c|c|}
    \hline
        	&	Sim.	&		\multicolumn{3}{c|}{Avg cost}			&			\multicolumn{3}{c|}{Avg step time (secs)}			&			\multicolumn{3}{c|}{Success}			\\ \hline
Domain	&		&	Rnd	&	$\pi_{h_{add}}(s)$	&	$\pi_{h_{add}}(s,B)$	&	Rnd	&	$\pi_{h_{add}}(s)$	&	$\pi_{h_{add}}(s,B)$	&	Rnd	&	$\pi_{h_{add}}(s)$	&	$\pi_{h_{add}}(s,B)$	\\ \hline\hline
doors 5	&	1500	&	17.3	&	\textbf{13.7}	&	14.4	&	6.552	&	0.403	&	1.86	&	100\%	&	100\%	&	100\%	\\ \hline
blocks 4	&	1500	&	4.9	&	4.45	&	4.8	&	0.33	&	0.13	&	0.37	&	100\%	&	100\%	&	100\%	\\ \hline
localize 3	&	1500	&	14.75	&	\textbf{10.35}	\textbf{}&	\textbf{10.95}	&	1.325	&	0.752	&	1.16	&	80\%	&	\textbf{100\%}	&	\textbf{100\%}	\\ \hline
MedPks 10	&	1500	&	\textbf{6.3}	&	7.25	&	7.45	&	4.029	&	4.594	&	3.388	&	100\%	&	100\%	&	100\%	\\ \hline
Unix 1	&	1500	&	7.35	&	\textbf{5.65}	&	6.4	&	2.551	&	0.274	&	1.322	&	100\%	&	100\%	&	100\%	\\ \hline
Wumpus 5	&	1500	&	39.47	&	33.352	&	\textbf{24.33}	&	10.829	&	3.491	&	4.909	&	85\%	&	85\%	&	\textbf{90\%}	\\ \hline
Wumpus 5	&	500	&	56.117	&	33.722	&	\textbf{22.05}	&	2.442	&	0.823	&	1.12	&	85\%	&	90\%	&	\textbf{100\%}	\\ \hline

    \end{tabular}
    \caption{Comparing rollout heuristics on various domains over 20 runs on each problem. Each number following the domain name is a specific instance of the domain, for example Wumpus5 means the grid is 5x5.}
    \label{tbl:Results}
\end{table}

Table~\ref{tbl:Results} presents the experiments results over the benchmarks, comparing the random (uniform) rollout policy (denoted Rnd), the $h_{add}$ heuristic using a single state ($\pi_{h_{add}}(s)$), and the $h_{add}$ heuristic over multiple states ($\pi_{h_{add}}(s,B)$). 

We begin by looking at the quality of the policy --- the average cost to the goal. As can be seen, the random rollout policy is best only in the MedPks domain, and close to best in unix. This is not too surprising, because in these two domains the best strategy is very simple, and random strategies easily stumble upon the goal. In blocks all methods achieved similar performance, because the optimal strategy is very short, and rollouts are less important. This domain has many possible actions, and hence a huge branching factor, making it difficult to scale up using POMCP.

On doors and localize, which require lengthier trajectories to reach the goal, but do not need long information-gathering efforts, the single state $h_{add}$ strategy operates very well. However, on Wumpus, where long sequences of actions are needed for information gathering, the multiple-state heuristic works best. We expected the results to be that way, although we expected more significant difference between the "smart" heuristics and the random rollout. 

With respect to the required time to run the simulations for a single decision, the results are mixed. While obviously the random strategy requires no time to compute the next action during a rollout, it often results in lengthy rollouts, which reduce this effect. The single state heuristic is almost always faster than the multi-state heuristic, but not by much.

\section{Conclusion}

In this paper we suggested to model goal POMDPs as stochastic contingent planning models, which allows us to use domain independent heuristics developed in the automated planning community to estimate the utility of belief states.
We implemented our domain independent heuristics into the rollout mechanism of POMCP --- a well known online POMDP planner that constructs a search tree to evaluate which action to take next. We provide an empirical evaluation showing how heuristics provide much leverage, especially in complex domains that require a long planning horizon, compared to the standard uniform rollout policy that is often used in POMCP.

For future research we intend to integrate our methods into other solvers, such as RTDP-BEL, or into point-based planners as a method to gather good belief points. We can also experiment with additional heuristics, other than the $h_{add}$ heuristic used in this paper.

\nocite{*}
\bibliographystyle{eptcs}
\bibliography{generic}

\begin{thebibliography}{10}
\providecommand{\bibitemdeclare}[2]{}
\providecommand{\surnamestart}{}
\providecommand{\surnameend}{}
\providecommand{\urlprefix}{Available at }
\providecommand{\url}[1]{\texttt{#1}}
\providecommand{\href}[2]{\texttt{#2}}
\providecommand{\urlalt}[2]{\href{#1}{#2}}
\providecommand{\doi}[1]{doi:\urlalt{https://doi.org/#1}{#1}}
\providecommand{\eprint}[1]{arXiv:\urlalt{https://arxiv.org/abs/#1}{#1}}
\providecommand{\bibinfo}[2]{#2}

\bibitemdeclare{inproceedings}{albore2009translation}
\bibitem{albore2009translation}
\bibinfo{author}{Alexandre \surnamestart Albore\surnameend},
  \bibinfo{author}{H{\'e}ctor \surnamestart Palacios\surnameend} \&
  \bibinfo{author}{Hector \surnamestart Geffner\surnameend}
  (\bibinfo{year}{2009}): \emph{\bibinfo{title}{A Translation-Based Approach to
  Contingent Planning.}}
\newblock In: {\slshape \bibinfo{booktitle}{IJCAI}}, \bibinfo{volume}{9}, pp.
  \bibinfo{pages}{1623--1628}, \doi{10.5555/1661445.1661706}.
\newblock \urlprefix\url{https://dl.acm.org/doi/10.5555/1661445.1661706}.

\bibitemdeclare{article}{bonet2001planning}
\bibitem{bonet2001planning}
\bibinfo{author}{Blai \surnamestart Bonet\surnameend} \&
  \bibinfo{author}{H{\'e}ctor \surnamestart Geffner\surnameend}
  (\bibinfo{year}{2001}): \emph{\bibinfo{title}{Planning as heuristic search}}.
\newblock {\slshape \bibinfo{journal}{Artificial Intelligence}}
  \bibinfo{volume}{129}(\bibinfo{number}{1-2}), pp. \bibinfo{pages}{5--33},
  \doi{10.1016/S0004-3702(01)00108-4}.

\bibitemdeclare{inproceedings}{bonet2009solving}
\bibitem{bonet2009solving}
\bibinfo{author}{Blai \surnamestart Bonet\surnameend} \&
  \bibinfo{author}{Hector \surnamestart Geffner\surnameend}
  (\bibinfo{year}{2009}): \emph{\bibinfo{title}{Solving POMDPs: RTDP-Bel vs.
  Point-based Algorithms}}.
\newblock In \bibinfo{editor}{Craig \surnamestart Boutilier\surnameend},
  editor: {\slshape \bibinfo{booktitle}{{IJCAI} 2009, Proceedings of the 21st
  International Joint Conference on Artificial Intelligence, Pasadena,
  California, USA, July 11-17, 2009}}, pp. \bibinfo{pages}{1641--1646},
  \doi{10.5555/1661445.1661709}.
\newblock \urlprefix\url{https://dl.acm.org/doi/10.5555/1661445.1661709}.

\bibitemdeclare{inproceedings}{bonet2011planning}
\bibitem{bonet2011planning}
\bibinfo{author}{Blai \surnamestart Bonet\surnameend} \&
  \bibinfo{author}{Hector \surnamestart Geffner\surnameend}
  (\bibinfo{year}{2011}): \emph{\bibinfo{title}{Planning under Partial
  Observability by Classical Replanning: Theory and Experiments}}.
\newblock In: {\slshape \bibinfo{booktitle}{{IJCAI} 2011, Proceedings of the
  22nd International Joint Conference on Artificial Intelligence, Barcelona,
  Catalonia, Spain, July 16-22, 2011}}, pp. \bibinfo{pages}{1936--1941},
  \doi{10.5591/978-1-57735-516-8/IJCAI11-324}.

\bibitemdeclare{article}{brafman2016online}
\bibitem{brafman2016online}
\bibinfo{author}{Ronen~I \surnamestart Brafman\surnameend} \&
  \bibinfo{author}{Guy \surnamestart Shani\surnameend} (\bibinfo{year}{2016}):
  \emph{\bibinfo{title}{Online belief tracking using regression for contingent
  planning}}.
\newblock {\slshape \bibinfo{journal}{Artificial Intelligence}}
  \bibinfo{volume}{241}, pp. \bibinfo{pages}{131--152},
  \doi{10.1016/j.artint.2016.08.005}.

\bibitemdeclare{book}{geffner2022concise}
\bibitem{geffner2022concise}
\bibinfo{author}{Hector \surnamestart Geffner\surnameend} \&
  \bibinfo{author}{Blai \surnamestart Bonet\surnameend} (\bibinfo{year}{2022}):
  \emph{\bibinfo{title}{A concise introduction to models and methods for
  automated planning}}.
\newblock \bibinfo{publisher}{Springer Nature}.

\bibitemdeclare{inproceedings}{helmert2009landmarks}
\bibitem{helmert2009landmarks}
\bibinfo{author}{Malte \surnamestart Helmert\surnameend} \&
  \bibinfo{author}{Carmel \surnamestart Domshlak\surnameend}
  (\bibinfo{year}{2009}): \emph{\bibinfo{title}{Landmarks, Critical Paths and
  Abstractions: What's the Difference Anyway?}}
\newblock In \bibinfo{editor}{Lubos \surnamestart Brim\surnameend},
  \bibinfo{editor}{Stefan \surnamestart Edelkamp\surnameend},
  \bibinfo{editor}{Eric~A. \surnamestart Hansen\surnameend} \&
  \bibinfo{editor}{Peter \surnamestart Sanders\surnameend}, editors: {\slshape
  \bibinfo{booktitle}{Graph Search Engineering, 29.11. - 04.12.2009}},
  {\slshape \bibinfo{series}{Dagstuhl Seminar Proceedings}}
  \bibinfo{volume}{09491}, \bibinfo{publisher}{Schloss Dagstuhl -
  Leibniz-Zentrum f{\"{u}}r Informatik, Germany},
  \doi{10.1609/icaps.v19i1.13370}.
\newblock \urlprefix\url{http://drops.dagstuhl.de/opus/volltexte/2010/2432/}.

\bibitemdeclare{article}{helmert2014merge}
\bibitem{helmert2014merge}
\bibinfo{author}{Malte \surnamestart Helmert\surnameend},
  \bibinfo{author}{Patrik \surnamestart Haslum\surnameend},
  \bibinfo{author}{J{\"o}rg \surnamestart Hoffmann\surnameend} \&
  \bibinfo{author}{Raz \surnamestart Nissim\surnameend} (\bibinfo{year}{2014}):
  \emph{\bibinfo{title}{Merge-and-shrink abstraction: A method for generating
  lower bounds in factored state spaces}}.
\newblock {\slshape \bibinfo{journal}{Journal of the ACM (JACM)}}
  \bibinfo{volume}{61}(\bibinfo{number}{3}), pp. \bibinfo{pages}{1--63},
  \doi{10.1145/2559951}.

\bibitemdeclare{article}{hff}
\bibitem{hff}
\bibinfo{author}{J.~\surnamestart Hoffmann\surnameend} \&
  \bibinfo{author}{B.~\surnamestart Nebel\surnameend} (\bibinfo{year}{2001}):
  \emph{\bibinfo{title}{The {FF} Planning System: Fast Plan Generation Through
  Heuristic Search}}.
\newblock {\slshape \bibinfo{journal}{JAIR}} \bibinfo{volume}{14}, pp.
  \bibinfo{pages}{253--302}, \doi{10.1613/jair.855}.

\bibitemdeclare{inproceedings}{hoffmann2005contingent}
\bibitem{hoffmann2005contingent}
\bibinfo{author}{J{\"{o}}rg \surnamestart Hoffmann\surnameend} \&
  \bibinfo{author}{Ronen~I. \surnamestart Brafman\surnameend}
  (\bibinfo{year}{2005}): \emph{\bibinfo{title}{Contingent Planning via
  Heuristic Forward Search witn Implicit Belief States}}.
\newblock In \bibinfo{editor}{Susanne \surnamestart Biundo\surnameend},
  \bibinfo{editor}{Karen~L. \surnamestart Myers\surnameend} \&
  \bibinfo{editor}{Kanna \surnamestart Rajan\surnameend}, editors: {\slshape
  \bibinfo{booktitle}{Proceedings of the Fifteenth International Conference on
  Automated Planning and Scheduling {(ICAPS} 2005), June 5-10 2005, Monterey,
  California, {USA}}}, \bibinfo{publisher}{{AAAI}}, pp.
  \bibinfo{pages}{71--80}, \doi{10.5555/3037062.3037072}.
\newblock
  \urlprefix\url{http://www.aaai.org/Library/ICAPS/2005/icaps05-008.php}.

\bibitemdeclare{inproceedings}{hoerger2019multilevel}
\bibitem{hoerger2019multilevel}
\bibinfo{author}{Marcus \surnamestart H{\"{o}}rger\surnameend},
  \bibinfo{author}{Hanna \surnamestart Kurniawati\surnameend} \&
  \bibinfo{author}{Alberto \surnamestart Elfes\surnameend}
  (\bibinfo{year}{2019}): \emph{\bibinfo{title}{Multilevel Monte-Carlo for
  Solving POMDPs Online}}.
\newblock In \bibinfo{editor}{Tamim \surnamestart Asfour\surnameend},
  \bibinfo{editor}{Eiichi \surnamestart Yoshida\surnameend},
  \bibinfo{editor}{Jaeheung \surnamestart Park\surnameend},
  \bibinfo{editor}{Henrik \surnamestart Christensen\surnameend} \&
  \bibinfo{editor}{Oussama \surnamestart Khatib\surnameend}, editors: {\slshape
  \bibinfo{booktitle}{Robotics Research - The 19th International Symposium
  {ISRR} 2019, Hanoi, Vietnam, October 6-10, 2019}}, {\slshape
  \bibinfo{series}{Springer Proceedings in Advanced
  Robotics}}~\bibinfo{volume}{20}, \bibinfo{publisher}{Springer}, pp.
  \bibinfo{pages}{174--190}, \doi{10.1007/978-3-030-95459-8\_11}.

\bibitemdeclare{article}{howard1966information}
\bibitem{howard1966information}
\bibinfo{author}{Ronald~A \surnamestart Howard\surnameend}
  (\bibinfo{year}{1966}): \emph{\bibinfo{title}{Information value theory}}.
\newblock {\slshape \bibinfo{journal}{IEEE Transactions on systems science and
  cybernetics}} \bibinfo{volume}{2}(\bibinfo{number}{1}), pp.
  \bibinfo{pages}{22--26}, \doi{10.1109/TSSC.1966.300074}.

\bibitemdeclare{inproceedings}{keller2012prost}
\bibitem{keller2012prost}
\bibinfo{author}{Thomas \surnamestart Keller\surnameend} \&
  \bibinfo{author}{Patrick \surnamestart Eyerich\surnameend}
  (\bibinfo{year}{2012}): \emph{\bibinfo{title}{{PROST:} Probabilistic Planning
  Based on {UCT}}}.
\newblock In \bibinfo{editor}{Lee \surnamestart McCluskey\surnameend},
  \bibinfo{editor}{Brian~Charles \surnamestart Williams\surnameend},
  \bibinfo{editor}{Jos{\'{e}}~Reinaldo \surnamestart Silva\surnameend} \&
  \bibinfo{editor}{Blai \surnamestart Bonet\surnameend}, editors: {\slshape
  \bibinfo{booktitle}{Proceedings of the Twenty-Second International Conference
  on Automated Planning and Scheduling, {ICAPS} 2012, Atibaia, S{\~{a}}o Paulo,
  Brazil, June 25-19, 2012}}, \bibinfo{publisher}{{AAAI}},
  \doi{10.1609/icaps.v22i1.13518}.
\newblock
  \urlprefix\url{http://www.aaai.org/ocs/index.php/ICAPS/ICAPS12/paper/view/4715}.

\bibitemdeclare{inproceedings}{keller2013trial}
\bibitem{keller2013trial}
\bibinfo{author}{Thomas \surnamestart Keller\surnameend} \&
  \bibinfo{author}{Malte \surnamestart Helmert\surnameend}
  (\bibinfo{year}{2013}): \emph{\bibinfo{title}{Trial-based heuristic tree
  search for finite horizon MDPs}}.
\newblock In: {\slshape \bibinfo{booktitle}{Proceedings of the International
  Conference on Automated Planning and Scheduling}}, \bibinfo{volume}{23}, pp.
  \bibinfo{pages}{135--143}, \doi{10.1609/icaps.v23i1.13557}.
\newblock
  \urlprefix\url{http://www.aaai.org/ocs/index.php/ICAPS/ICAPS13/paper/view/6026}.

\bibitemdeclare{inproceedings}{kim2019pomhdp}
\bibitem{kim2019pomhdp}
\bibinfo{author}{Sung-Kyun \surnamestart Kim\surnameend}, \bibinfo{author}{Oren
  \surnamestart Salzman\surnameend} \& \bibinfo{author}{Maxim \surnamestart
  Likhachev\surnameend} (\bibinfo{year}{2019}): \emph{\bibinfo{title}{POMHDP:
  Search-based belief space planning using multiple heuristics}}.
\newblock In: {\slshape \bibinfo{booktitle}{Proceedings of the International
  Conference on Automated Planning and Scheduling}}, \bibinfo{volume}{29}, pp.
  \bibinfo{pages}{734--744}, \doi{10.1609/icaps.v29i1.3542}.
\newblock
  \urlprefix\url{https://ojs.aaai.org/index.php/ICAPS/article/view/3542}.

\bibitemdeclare{article}{kurniawati2022partially}
\bibitem{kurniawati2022partially}
\bibinfo{author}{Hanna \surnamestart Kurniawati\surnameend}
  (\bibinfo{year}{2022}): \emph{\bibinfo{title}{Partially observable markov
  decision processes and robotics}}.
\newblock {\slshape \bibinfo{journal}{Annual Review of Control, Robotics, and
  Autonomous Systems}} \bibinfo{volume}{5}, pp. \bibinfo{pages}{253--277},
  \doi{10.1146/annurev-control-042920-092451}.

\bibitemdeclare{inproceedings}{kurniawati2016online}
\bibitem{kurniawati2016online}
\bibinfo{author}{Hanna \surnamestart Kurniawati\surnameend} \&
  \bibinfo{author}{Vinay \surnamestart Yadav\surnameend}
  (\bibinfo{year}{2016}): \emph{\bibinfo{title}{An online POMDP solver for
  uncertainty planning in dynamic environment}}.
\newblock In: {\slshape \bibinfo{booktitle}{Robotics Research: The 16th
  International Symposium ISRR}}, \bibinfo{organization}{Springer}, pp.
  \bibinfo{pages}{611--629}, \doi{10.1007/978-3-319-28872-7_35}.

\bibitemdeclare{article}{luo2019importance}
\bibitem{luo2019importance}
\bibinfo{author}{Yuanfu \surnamestart Luo\surnameend}, \bibinfo{author}{Haoyu
  \surnamestart Bai\surnameend}, \bibinfo{author}{David \surnamestart
  Hsu\surnameend} \& \bibinfo{author}{Wee~Sun \surnamestart Lee\surnameend}
  (\bibinfo{year}{2019}): \emph{\bibinfo{title}{Importance sampling for online
  planning under uncertainty}}.
\newblock {\slshape \bibinfo{journal}{The International Journal of Robotics
  Research}} \bibinfo{volume}{38}(\bibinfo{number}{2-3}), pp.
  \bibinfo{pages}{162--181}, \doi{10.1177/0278364918780322}.

\bibitemdeclare{inproceedings}{richter2008landmarks}
\bibitem{richter2008landmarks}
\bibinfo{author}{Silvia \surnamestart Richter\surnameend},
  \bibinfo{author}{Malte \surnamestart Helmert\surnameend} \&
  \bibinfo{author}{Matthias \surnamestart Westphal\surnameend}
  (\bibinfo{year}{2008}): \emph{\bibinfo{title}{Landmarks Revisited.}}
\newblock In: {\slshape \bibinfo{booktitle}{AAAI}}, \bibinfo{volume}{8}, pp.
  \bibinfo{pages}{975--982}, \doi{10.2307/j.ctt1zxsjcs}.
\newblock \urlprefix\url{http://www.aaai.org/Library/AAAI/2008/aaai08-155.php}.

\bibitemdeclare{incollection}{rintanen2008regression}
\bibitem{rintanen2008regression}
\bibinfo{author}{Jussi \surnamestart Rintanen\surnameend}
  (\bibinfo{year}{2008}): \emph{\bibinfo{title}{Regression for classical and
  nondeterministic planning}}.
\newblock In: {\slshape \bibinfo{booktitle}{ECAI 2008}},
  \bibinfo{publisher}{IOS Press}, pp. \bibinfo{pages}{568--572},
  \doi{10.3233/978-1-58603-891-5-568}.

\bibitemdeclare{article}{ross2008online}
\bibitem{ross2008online}
\bibinfo{author}{St{\'e}phane \surnamestart Ross\surnameend},
  \bibinfo{author}{Joelle \surnamestart Pineau\surnameend},
  \bibinfo{author}{S{\'e}bastien \surnamestart Paquet\surnameend} \&
  \bibinfo{author}{Brahim \surnamestart Chaib-Draa\surnameend}
  (\bibinfo{year}{2008}): \emph{\bibinfo{title}{Online planning algorithms for
  POMDPs}}.
\newblock {\slshape \bibinfo{journal}{Journal of Artificial Intelligence
  Research}} \bibinfo{volume}{32}, pp. \bibinfo{pages}{663--704},
  \doi{10.1613/jair.2567}.

\bibitemdeclare{inproceedings}{saborio2019planning}
\bibitem{saborio2019planning}
\bibinfo{author}{Juan~Carlos \surnamestart Sabor{\'\i}o\surnameend} \&
  \bibinfo{author}{Joachim \surnamestart Hertzberg\surnameend}
  (\bibinfo{year}{2019}): \emph{\bibinfo{title}{Planning Under Uncertainty
  Through Goal-Driven Action Selection}}.
\newblock In: {\slshape \bibinfo{booktitle}{Agents and Artificial Intelligence:
  10th International Conference, ICAART 2018, Funchal, Madeira, Portugal,
  January 16--18, 2018, Revised Selected Papers 10}},
  \bibinfo{organization}{Springer}, pp. \bibinfo{pages}{182--201},
  \doi{10.1007/978-3-030-05453-3_9}.

\bibitemdeclare{inproceedings}{saborio2020efficient}
\bibitem{saborio2020efficient}
\bibinfo{author}{Juan~Carlos \surnamestart Sabor{\'\i}o\surnameend} \&
  \bibinfo{author}{Joachim \surnamestart Hertzberg\surnameend}
  (\bibinfo{year}{2020}): \emph{\bibinfo{title}{Efficient planning under
  uncertainty with incremental refinement}}.
\newblock In: {\slshape \bibinfo{booktitle}{Uncertainty in Artificial
  Intelligence}}, \bibinfo{organization}{PMLR}, pp. \bibinfo{pages}{303--312},
  \doi{10.1109/TSMC.1987.4309045}.

\bibitemdeclare{inproceedings}{shani2011replanning}
\bibitem{shani2011replanning}
\bibinfo{author}{Guy \surnamestart Shani\surnameend} \&
  \bibinfo{author}{Ronen~I \surnamestart Brafman\surnameend}
  (\bibinfo{year}{2011}): \emph{\bibinfo{title}{Replanning in domains with
  partial information and sensing actions}}.
\newblock In: {\slshape \bibinfo{booktitle}{IJCAI}}, \bibinfo{volume}{2011},
  \bibinfo{organization}{Citeseer}, pp. \bibinfo{pages}{2021--2026},
  \doi{10.5591/978-1-57735-516-8/IJCAI11-337}.

\bibitemdeclare{article}{shani2013survey}
\bibitem{shani2013survey}
\bibinfo{author}{Guy \surnamestart Shani\surnameend}, \bibinfo{author}{Joelle
  \surnamestart Pineau\surnameend} \& \bibinfo{author}{Robert \surnamestart
  Kaplow\surnameend} (\bibinfo{year}{2013}): \emph{\bibinfo{title}{A survey of
  point-based POMDP solvers}}.
\newblock {\slshape \bibinfo{journal}{Autonomous Agents and Multi-Agent
  Systems}} \bibinfo{volume}{27}, pp. \bibinfo{pages}{1--51},
  \doi{10.1007/s10458-012-9200-2}.

\bibitemdeclare{inproceedings}{shen2019guiding}
\bibitem{shen2019guiding}
\bibinfo{author}{William \surnamestart Shen\surnameend},
  \bibinfo{author}{Felipe \surnamestart Trevizan\surnameend},
  \bibinfo{author}{Sam \surnamestart Toyer\surnameend}, \bibinfo{author}{Sylvie
  \surnamestart Thi{\'e}baux\surnameend} \& \bibinfo{author}{Lexing
  \surnamestart Xie\surnameend} (\bibinfo{year}{2019}):
  \emph{\bibinfo{title}{Guiding search with generalized policies for
  probabilistic planning}}.
\newblock In: {\slshape \bibinfo{booktitle}{Proceedings of the International
  Symposium on Combinatorial Search}}, \bibinfo{volume}{10}, pp.
  \bibinfo{pages}{97--105}, \doi{10.1609/socs.v10i1.18507}.

\bibitemdeclare{article}{silver2010monte}
\bibitem{silver2010monte}
\bibinfo{author}{David \surnamestart Silver\surnameend} \&
  \bibinfo{author}{Joel \surnamestart Veness\surnameend}
  (\bibinfo{year}{2010}): \emph{\bibinfo{title}{Monte-Carlo planning in large
  POMDPs}}.
\newblock {\slshape \bibinfo{journal}{Advances in neural information processing
  systems}} \bibinfo{volume}{23}, \doi{10.5555/2997046.2997137}.

\bibitemdeclare{article}{smallwood1973optimal}
\bibitem{smallwood1973optimal}
\bibinfo{author}{Richard~D \surnamestart Smallwood\surnameend} \&
  \bibinfo{author}{Edward~J \surnamestart Sondik\surnameend}
  (\bibinfo{year}{1973}): \emph{\bibinfo{title}{The optimal control of
  partially observable Markov processes over a finite horizon}}.
\newblock {\slshape \bibinfo{journal}{Operations research}}
  \bibinfo{volume}{21}(\bibinfo{number}{5}), pp. \bibinfo{pages}{1071--1088},
  \doi{10.1287/opre.21.5.1071}.

\bibitemdeclare{article}{somani2013despot}
\bibitem{somani2013despot}
\bibinfo{author}{Adhiraj \surnamestart Somani\surnameend}, \bibinfo{author}{Nan
  \surnamestart Ye\surnameend}, \bibinfo{author}{David \surnamestart
  Hsu\surnameend} \& \bibinfo{author}{Wee~Sun \surnamestart Lee\surnameend}
  (\bibinfo{year}{2013}): \emph{\bibinfo{title}{DESPOT: Online POMDP planning
  with regularization}}.
\newblock {\slshape \bibinfo{journal}{Advances in neural information processing
  systems}} \bibinfo{volume}{26}, \doi{10.1613/jair.5328}.

\bibitemdeclare{inproceedings}{sunberg2018online}
\bibitem{sunberg2018online}
\bibinfo{author}{Zachary \surnamestart Sunberg\surnameend} \&
  \bibinfo{author}{Mykel \surnamestart Kochenderfer\surnameend}
  (\bibinfo{year}{2018}): \emph{\bibinfo{title}{Online algorithms for POMDPs
  with continuous state, action, and observation spaces}}.
\newblock In: {\slshape \bibinfo{booktitle}{Proceedings of the International
  Conference on Automated Planning and Scheduling}}, \bibinfo{volume}{28}, pp.
  \bibinfo{pages}{259--263}, \doi{10.48550/arXiv.1709.06196}.

\bibitemdeclare{inproceedings}{xiao2019online}
\bibitem{xiao2019online}
\bibinfo{author}{Yuchen \surnamestart Xiao\surnameend}, \bibinfo{author}{Sammie
  \surnamestart Katt\surnameend}, \bibinfo{author}{Andreas \surnamestart ten
  Pas\surnameend}, \bibinfo{author}{Shengjian \surnamestart Chen\surnameend} \&
  \bibinfo{author}{Christopher \surnamestart Amato\surnameend}
  (\bibinfo{year}{2019}): \emph{\bibinfo{title}{Online planning for target
  object search in clutter under partial observability}}.
\newblock In: {\slshape \bibinfo{booktitle}{2019 International Conference on
  Robotics and Automation (ICRA)}}, \bibinfo{organization}{IEEE}, pp.
  \bibinfo{pages}{8241--8247}, \doi{10.1109/ICRA.2019.8793494}.

\bibitemdeclare{article}{ye2017despot}
\bibitem{ye2017despot}
\bibinfo{author}{Nan \surnamestart Ye\surnameend}, \bibinfo{author}{Adhiraj
  \surnamestart Somani\surnameend}, \bibinfo{author}{David \surnamestart
  Hsu\surnameend} \& \bibinfo{author}{Wee~Sun \surnamestart Lee\surnameend}
  (\bibinfo{year}{2017}): \emph{\bibinfo{title}{DESPOT: Online POMDP planning
  with regularization}}.
\newblock {\slshape \bibinfo{journal}{Journal of Artificial Intelligence
  Research}} \bibinfo{volume}{58}, pp. \bibinfo{pages}{231--266},
  \doi{10.1613/jair.5328}.

\end{thebibliography}
\end{document}